\documentclass[12pt]{article}
\usepackage[utf8]{inputenc}
\usepackage[T1]{fontenc}
\usepackage{graphicx}
\usepackage{url}
\usepackage{amssymb}
\usepackage{amsmath}
\usepackage{cite}
\usepackage{caption}
\usepackage{subcaption}
\usepackage{float}
\usepackage{dirtree}
\usepackage[edges]{forest}
\usepackage{fontawesome}
\usepackage[utf8]{inputenc}
\usepackage{parskip}
\usepackage{soul}
\usepackage{xcolor}
\usepackage{tipa}
\usepackage{fancyhdr}
\usepackage[breaklinks]{hyperref}
\pdfoutput = 1
\author{Nigar Alishzade\thanks{nigar.alishzada@ufaz.az}\\
\textit{Azerbaijan State Oil and Industry University, Baku, Azerbaijan} \\
\textit{French-Azerbaijani University, Baku, Azerbaijan}\and
Jamaladdin Hasanov\thanks{jhasanov@ada.edu.az}\\
\textit{ADA University, Baku, Azerbaijan}}
\date{}

\DeclareUnicodeCharacter{018F}{}
\DeclareUnicodeCharacter{0259}{}

\begin{document}

\title{AzSLD: Azerbaijani Sign Language Dataset for Fingerspelling, Word, and Sentence Translation with Baseline Software}
\maketitle

\begin{abstract}
Sign language processing technology development relies on extensive and reliable datasets, instructions, and ethical guidelines. We present a comprehensive Azerbaijani Sign Language Dataset (AzSLD) collected from diverse sign language users and linguistic parameters to facilitate advancements in sign recognition and translation systems and support the local sign language community. The dataset was created within the framework of a vision-based AzSL translation project. This study introduces the dataset as a summary of the fingerspelling alphabet and sentence- and word-level sign language datasets. The dataset was collected from signers of different ages, genders, and signing styles, with videos recorded from two camera angles to capture each sign in full detail. This approach ensures robust training and evaluation of gesture recognition models. AzSLD contains 30,000 videos, each carefully annotated with accurate sign labels and corresponding linguistic translations. The dataset is accompanied by technical documentation and source code to facilitate its use in training and testing. This dataset offers a valuable resource of labeled data for researchers and developers working on sign language recognition, translation, or synthesis. Ethical guidelines were strictly followed throughout the project, with all participants providing informed consent for collecting, publishing, and using the data.
\end{abstract}


\section{Introduction}

Sign language technology concerns all the non-verbal people worldwide who use a certain local sign language, making it an important computer vision task. Sign Languages (SLs) are composed of manual and non-manual components through which complex semantics can be conveyed. Sign language technologies help bridge the communication gap between signers and non-signers and facilitate the interaction between the Deaf and hard-of-hearing (DHH) community and technology. For instance, sign language technologies can empower DHH content creators to produce and edit sign language videos more easily.

Understanding sign language videos has been tackled from a range of tasks, including sign language recognition (SLR) \cite{KASAPBASI2022100048, DU2022115, MUSTHAFA2022504, KUMAR2022143, cbam-resnet, DAS2023118914, Katoch2022, Rastgoo2021, Ma2022, IMRAN2021107021}, sign language translation (SLT) \cite{ZHANG2023110526, DECASTRO2023119394, skeleton, contrastive, signer-adaptive, Li2023, Ananthanarayana2021}, and identification \cite{gebre-etal-2014-unsupervised}. In particular, end-to-end SLT systems can overcome communication barriers for deaf and hard-of-hearing individuals, contributing to their social integration.

SLR covers fingerspelling recognition and isolated sign classification as it is limited by the available computer vision technologies. SLT is an interdisciplinary research domain involving computer vision, machine translation, and linguistics \cite{MT_SOTA}. SLT requires large amounts of high-quality labeled data, which can be challenging to collect. In this work, we present the AzSL dataset, which is mainly concerned with SL data representation regarding the needs of SLR and SLT tasks.

According to \cite{bestpractice} and \cite{handbook}, collecting a sign language dataset for translation involves recording videos of signers signing in the target sign language along with annotations of the corresponding spoken language translations. These annotations could be in the form of labels that assign each sign a word from the spoken language that most appropriately describes the meaning of the sign. In addition, annotation metadata can include information about the signer, such as age, sex, and signing proficiency. It is crucial to ensure that the data set is diverse and representative of the target sign language community, including signers of different ages and sexes with slightly different signing speeds.

The authors of \cite{ASLLR} introduced the ASL Corpora and Sign Bank, which includes linguistic annotations for lexical signs and fingerspelling. This resource has been utilized in various research areas in linguistics and sign language recognition from videos. According to the author in \cite{Crasborn2022}, the creation of sign language datasets has had a significant impact on sign language research. It provides access to a comprehensive record of the language, catering to its diversity, which enables researchers to delve deeper into the study of sign languages. This has led to the development of new insights into their structure, usage, and evolution. Moreover, the availability of annotated corpora has paved the way for the advancement of language technologies, such as machine translation, which are becoming increasingly crucial.

The collection of sign language datasets for low-resource sign languages is of paramount importance, as these languages often have limited digital resources. Such datasets of low-resource sign languages lays the groundwork for fostering linguistic inclusivity and technological empowerment among underrepresented DHH populations.

In this paper, we suggest the AzSL Dataset as the first resource for Azerbaijani Sign Language with AI purposes, a large video set lasting approximately 65 hours combined with annotation data for training. We present information regarding the data collection and annotation of potential data users of the SL technology research community contributors in a clear and accessible manner. AzSLD is made available under the Creative Commons Attribution 4.0 International License on Zenodo repository \cite{azsld}. Data loaders for utilizing this dataset are publicly accessible through the GitHub repository \cite{AzSLdataloader}.

\section{Related work}

The quality of the training data plays a vital role in achieving accurate and reliable results for sign language alphabet and word-level recognition, as well as end-to-end sign language translation tasks \cite{us_deaf_community}. To ensure the quality of the training data, various metrics can be used to evaluate the annotations, including lexical correctness, completeness, consistency, diversity, and representativeness. 

To address the limitations of existing sign language datasets, researchers have been actively involved in collecting and annotating comprehensive datasets. In \cite{kopf-etal-2022-sign} authors provide an extensive overview of linguistic resources for sign languages worldwide. In Table \ref{tab:related_datasets} we present a list of statistics for publicly available sign language datasets, highlighting their key characteristics.

\begin{table}[h]
\resizebox{\textwidth}{!}{
\begin{tabular}{|l|l|l|l|l|l|l|l|}
\hline
Name & Language & \#Videos & \#Hours & \#Signers & Annotation type & Source & Loader \\ \hline
ASL Citizen \cite{desai2023asl} & ASL & 83,399 & - & 52 & Word-level & Crowd & \\
PopSign ASL v1.0 \cite{starner2023popsign} & ASL & 214,326 & - & 47 & Word-level & Crowd & \\
MyWSL2023 \cite{mywsl2023} & Malaysian & - & - & 5 & Word-level & Lab & \\
Youtube-ASL \cite{youtubeasl} & ASL & 60,000 & 984 & >2519 & Sentence-level & Web & \\
Alabib-65 \cite{alabib-65} & Algerian & 6,000 & - & 41 & Word-level & Lab & \\
SLOVO \cite{slovo} & Russian SL & 1,000 & 9.2 & 194 & Word-level & Lab and web & \checkmark \\
AfriSign \cite{afrisign} & various & 20,000 & 152 & - & Sentence-level & Web & \\
LSA-T \cite{lsa-t} & Spanish & 14,880 & 21.78 & 103 & Sentence-level & Real life & \\
OpenASL \cite{openasl} & ASL & 32,000 & 288 & 200 & Sentence-level & Web & \\
CSL-daily \cite{csldaily} & Chenese SL & 2,000 & 23 & 10 & Word-level and sentence-level & Lab & \\
FluentSigners-50 \cite{fluentsigners} & Kazakh-Russian & - & - & 50 & Sentence-level & Lab & \\
LSFB Datasets \cite{lsfb} & French-Belgian SL & 90 & 50 & 100 & Word-level and sentence-level & Lab & \checkmark \\
How2Sign \cite{how2sign} & ASL & 16,000 & 80 & 11 & Sentence-level & Lab & \checkmark \\
BosphorusSign22k \cite{bosphorussign22k} & Turkish SL & 22,000 & 19 & 6 & Word-level & Lab & \\
RWTH-PHOENIX-2014T \cite{phoenix} & German & 3,000 & 11 & 9 & Word-level and sentence-level & TV & \\
BOBSL \cite{bobsl} & British & 77,000 & 1447 & 39 & Sentence-level & TV & \\
MS-ASL \cite{ms-asl} & ASL & 25,000 & 24 & 200 & Sentence-level & Lab & \\
WLASL \cite{wl-asl} & ASL & 21,000 & 14 & 100 & Word-level & Lab & \checkmark \\
AUTSL \cite{autsl} & Turkish & 38,000 & - & 43 & Word-level & Lab & \\
KETI \cite{keti} & Korean & 419,000 & 28 & 14 & Sentence-level & Lab & \\
Polytropon Parallel Corpus \cite{polytropon} & Greek & 3,600 & - & 1 & Word-level and sentence-level & Lab & \\
K-RSL \cite{k-rsl} & Kazakh-Russian & 28K & - & 10 & Word-level & Lab & \\ \hline
AzSL (our) \cite{azsld} & Azerbaijani SL & 30,312 & 65.2 & 43 & Word-level and sentence-level & Lab & \checkmark \\ \hline
\end{tabular}
}
\caption{Summary statistics for related Sign Language datasets}
\label{tab:related_datasets}
\end{table}

The number of videos, signers, and variety of setups have a positive impact on the efficiency of training a custom pose estimation model using a sign language dataset. These factors indicate the diversity and richness of the dataset, which are crucial for training accurate and robust deep learning models. Fine-tuning the frozen layers of pre-trained human-pose models can be a viable approach when the length of videos in the dataset is insufficient \cite{openhands}. In this case, the pre-trained models convey the human pose features and patterns from the previous extensive training made on large datasets applicable to our domain \cite{survey2022}. By customizing and adapting these pre-trained models for the sign language dataset, the model can adapt to the specific characteristics and nuances of sign language gestures, improving its accuracy and performance.

Some of SL datasets lack annotation granularity \cite{acl-2021-association}. By including both word- and sentence-level annotations in a sign language dataset, researchers can better understand the learning process within the model layers of an end-to-end sign language translation pipeline \cite{Aloysius2021}. This will enable the analysis of the model’s process and interpretation of the individual signs (tokens) as well as bring a notion of how they integrate and generate coherent translations at the sentence level.

Each sign language dataset presents an opportunity to develop assistive sign language technologies tailored to specific communities, as well as to create new benchmarking resources for the research community. Adaptable and well-documented data loaders for these datasets enhance their utility by providing researchers with standardized tools for data access and analysis, thereby facilitating comparative studies and advancing the field. In this study, we address both of these aspects.

\section{AzSL Dataset}

\subsection{Dataset components}

Including both fingerspelling and large-lexicon signs, the dataset becomes more diverse, representative, and valuable for sign language research and development. In addition, the inclusion of both approaches facilitated the use of the dataset by recognition models with different complexities.

In AzSLD, we have added labels for isolated fingerspelling and word-level videos, and offer time-aligned Azerbaijani sentence-level translations for sentence videos. Thus, we have three main components of AzSLD:
\begin{itemize}
    \item AzSLD Fingerspelling
    \item AzSLD Words
    \item AzSLD Sentences
\end{itemize}

\subsubsection*{AzSLD Fingerspelling}

Fingerspelling is a subset of sign language that spells words by representing each letter of a written alphabet with specific hand shapes and movements. In the Azerbaijani alphabet, there are 32 letters and 32 different hand signs used in fingerspelling to represent each of these letters. Azerbaijani fingerspelling comprises of 24 static and 8 dynamic gestures. In the dataset, samples for the staticly expressed letters collected as images, and samples for dynamically expressed letteres collected as videos. 

\begin{table}[H]
\centering
\resizebox{\textwidth}{!}{
\begin{tabular}{|c|c|c|c|c|c|c|c|c|c|c|c|c|c|c|c|}
\hline
\textbf{a}   & \textbf{b}   & \textbf{c}   & \c{c}   & \textbf{d}   & \textbf{e}   & \textbf{\textschwa}   & \textbf{f}   & \textbf{g}   & \u{g}   & \textbf{x}   & \textbf{h}   & \textbf{\textiota}   & \textbf{i}   & \textbf{j}   & \textbf{k}   \\ \hline
800 & 770 & 661 & 385 & 450 & 423 & 434 & 382 & 459 & 398 & 396 & 430 & 403 & 366 & 514 & 464 \\ \hline
\textbf{q}   & \textbf{l}   & \textbf{m}   & \textbf{n}   & \textbf{o}   & \"{o}   & \textbf{p}   & \textbf{r}   & \textbf{s}   & \c{s}   & \textbf{t}   & \textbf{u}   & \"{u}   & \textbf{v}   & \textbf{y}   & \textbf{z}   \\ \hline
353 & 399 & 397 & 385 & 420 & 486 & 418 & 415 & 394 & 378 & 421 & 421 & 368 & 493 & 478 & 492 \\ \hline
\end{tabular}
}
\caption{Number of samples collected for each letter}
\label{tab:count_fingerspelling}
\vspace{-10mm}
\end{table}

To ensure an effective image collection process and make it accessible to a wide range of users, we developed \textit{JestDiliBot}, a Telegram bot that employs an interactive interface for data collection. Users are prompted to choose a letter and are then presented with a sample image or video corresponding to their selection. They were encouraged to capture and submit similar data. The fingerspelling data samples were collected from volunteers who were not native signers. The collected samples were checked and validated by the team of AzSL native signers. More than 200 volunteers contributed to the AzSL fingerspelling dataset. After validation there reported 10,864 images for static letters, and 3,587 videos for dynamic letters. The distributions of the images and videos for the letters are listed in Table \ref{tab:count_fingerspelling}. The collection of the fingerspelling dataset and the demonstration of a word recognition system are described in detail in \cite{Hasanov}.

\subsubsection*{AzSLD Words}

AzSLD Words comprises 100 classes of word videos, representing the top 100 most frequent words observed within the sentences of the AzSLD Sentences dataset. The number of samples in the word-level dataset is highly imbalanced across classes. Therefore, we recommend using class weights for accuracy measurement in word-level recognition models. The recording setup for word-level videos is the same as that used for sentence-level videos. Folder names indicate the ground truth labels for the ease of word-level model evaluation.

\subsubsection*{AzSLD Sentences}

The strategy used for the fingerspelling dataset is not applicable to the collection of the sentence dataset; the interpretation of each sentence and demonstration of the signs need to be moderated. AzSLD Sentences comprises signing videos for 500 sentences, collected using frontal and side-view cameras. Each dataset folder contains time-aligned sentence annotation files and two subfolders for the recordings from the two cameras. Ground truth annotations of sentences for each class were added in a separate file. The videos were performed by 18 to 25 different signers, with a slight imbalance among them. 

A custom setup was used for filming, with a white background and a fixed seating distance from the camera. RGB videos were collected in consideration of the results for human-action recognition studies with depth cameras \cite{Shafizadegan2024}. This type of setup makes a seamless application of transfer learning if a model has been pre-trained with an RGB or depth videos of a human-action dataset. 

\begin{figure}[H]
\centering
\includegraphics[width=\textwidth]{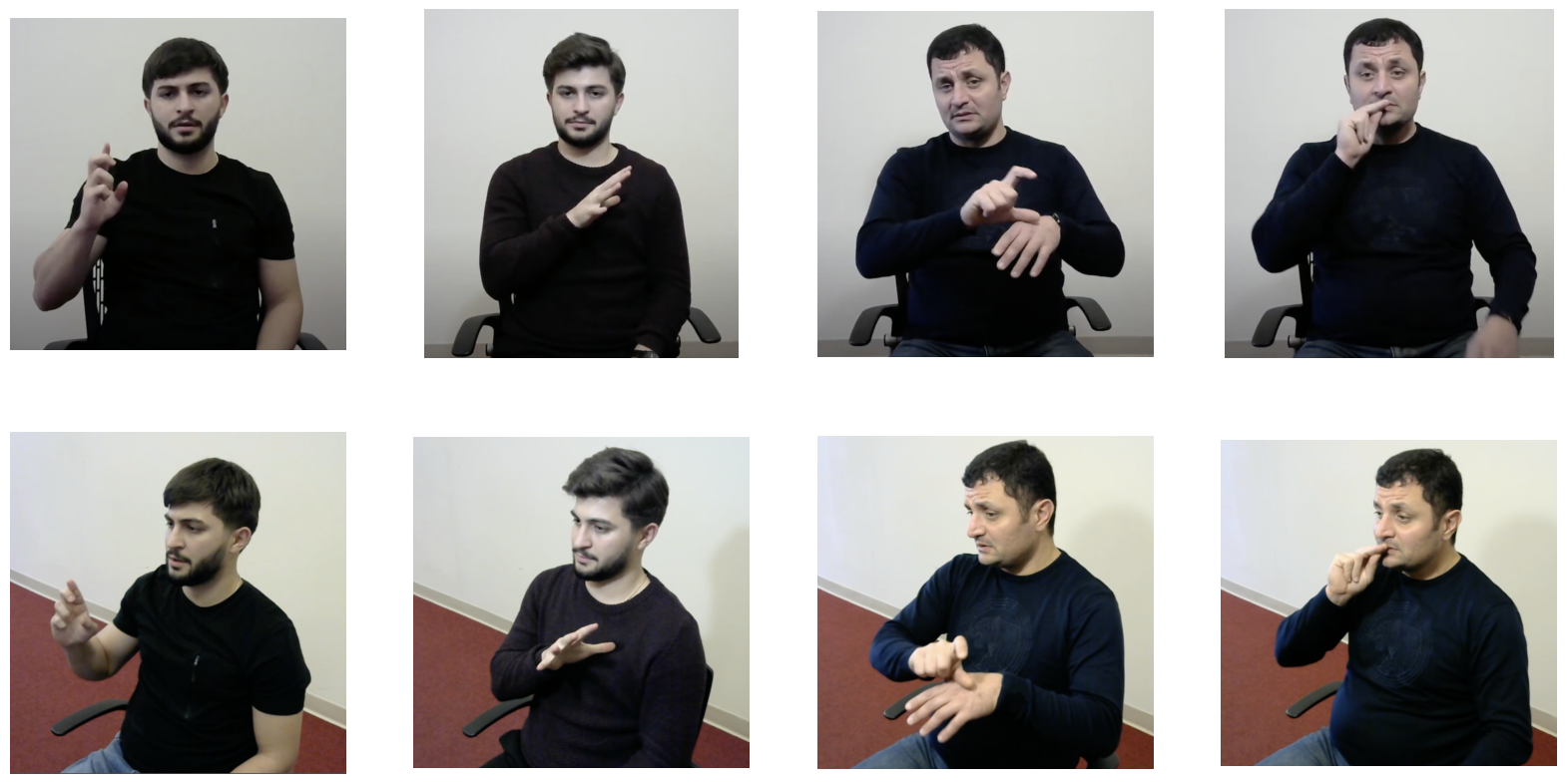}
\caption{Examples of the video screens captured by the frontal and the side view cameras}
\label{fig:front_side_cams}
\end{figure}

To obtain more information from the collected data, it was decided to capture the video using two cameras: one mounted on the monitor, looking straight at the actor, and another mounted on the top-left side. These two views are considered complimentary - those signs that are not visible from the front view could be noticed in the side view and vice versa (e.g., which finger has been folded or what the mouth mactivity when a hand accidentally covers it). Fig. \ref{fig:front_side_cams} demonstrates examples of screens from both cameras for 4 different signs. Video files from the two cameras have identical names and are located in the 'Cam1' and 'Cam2' folders, respectively in the sentence-level dataset.

Involving members of the sign language community in the dataset collection effort helps to ensure that the collected data are relevant and beneficial to the community. The AzSL dataset was annotated by fluent AzSL users to ensure its accuracy and reliability. 

For the labeling of the individual signs, the video files were uploaded to the Supervisely tool \cite{supervisely}, where each sign is associated with the word that corresponds to the start and end frames in the video.





\parbox[t]{\linewidth}{\subsection{Context, linguistic analysis, and annotations in AzSLD Sentences}}

Designing a national sign language dataset begins with a thorough understanding of the linguistic and cultural context of the target sign language \cite{Hochgesang2022}. Moreover, contextual information plays a crucial role in sign language understanding and translation. The first AzSL dataset provided in this paper was collaboratively developed by 42 DHH individuals who are native users of AzSLD, one CODA (Child Of Deaf Adults) person, Azerbaijani Sign Language interpreters, and social service workers specializing in assisting non-verbal individuals. The primary objective of this collaboration was to ensure that the dataset encompasses the entire glossary commonly used by individuals with hearing impairments in social service centers.

Social service agents composed and suggested sentences and short texts, including common communication phrases and question-and-answer exchanges. These materials were then reviewed and elaborated upon by members of the deaf community. Ultimately, 500 sentences were selected for inclusion in the dataset collection. By involving these stakeholders, the dataset captured a comprehensive range of relevant and meaningful signs and expressions within the context of social service centers. This is the most important field where the Deaf community needs assistance with communication. However, the scope of the dataset can be broadened in the future. 

To enhance usability and accessibility, the sentences were simplified. Simplifying the language used in the dataset makes it more inclusive and caters to a wider range of users, regardless of their sign language proficiency level.
Each sentence was recorded by at least 18 of the 40 participants involved in the project. The involvement of multiple signers ensured that the dataset captured a wide range of signing styles, signing speed variations, and nuances. This approach also accounts for individual differences in signing preferences and regional variations within Azerbaijani Sign Language.

\begin{figure}[H]
\centering
\includegraphics[width=\textwidth]{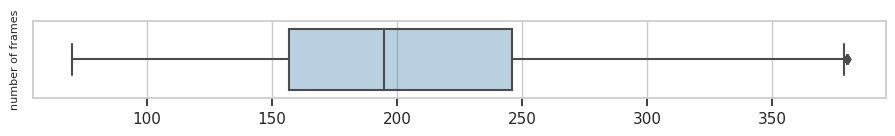}
\caption{Total frame numbers of large-lexicon signing videos}
\label{fig-frames}
\end{figure}

Sign languages have specific linguistic rules that do not directly reflect spoken language \cite{SLCompendium, kopf_maria_2021_9561}. Similar to the problem of translation between spoken languages, which opposes two sequences of variable length and tokens, sign language translation requires an understanding of the meaning and order of signs in a video \cite{Camgoz2018}. In an isolated sentence-level annotation, each distinct label, such as a word, number, or letter, is annotated with respect to the corresponding frame number in the video. This annotation type provides a broader context for feature extraction and allows the analysis of sentence-level features and structures. The total length of the video varied according to the number of words performed in the video. \ref{fig-frames} visualizes the variability of the length of videos in frame numbers.

\begin{figure}[H]
\centering
\includegraphics[width=\textwidth]{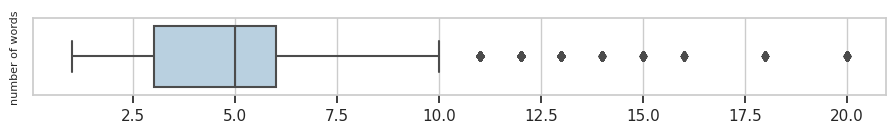}
\caption{Variety of word counts in large-lexicon signing videos}
\label{fig-words}
\end{figure}

\begin{figure}[H]
\centering
\includegraphics[width=\linewidth]{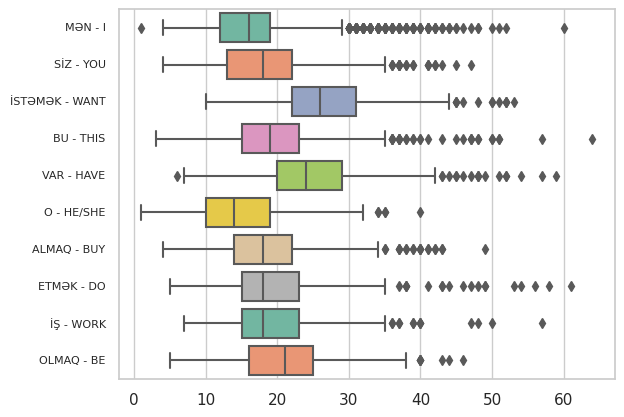}
\caption{Frame numbers of top 10 most frequent labels}
\label{top_10_boxplots}
\end{figure}

Within the 500 sentences of the dataset, there are 687 unique labels, some of which recur in different sentences. \ref{fig-words} shows the overall number of labels within sentences. Included sentences cover various contexts—including everyday conversations, directions, time and date expressions, purchasing items, medical inquiries, social service interactions, numbers, dates, and letters—to address various communication needs that individuals with hearing impairments might encounter. The location variation for the top 10 labels within 500 sentences is shown in Fig. \ref{top_10_boxplots}. 

Meticulous frame number alignment in video annotation is essential for synchronizing signs with their corresponding glosses or spoken words. By precisely mapping each sign to specific frames, annotators can accurately define the temporal boundaries of signs. This alignment enhances the analysis and understanding of sign language content and is crucial for developing effective sign language recognition and translation systems, as it captures the nuances and timing of each sign \cite{Hou2022}.

\begin{figure}[H]
\centering
\includegraphics[width=\textwidth]{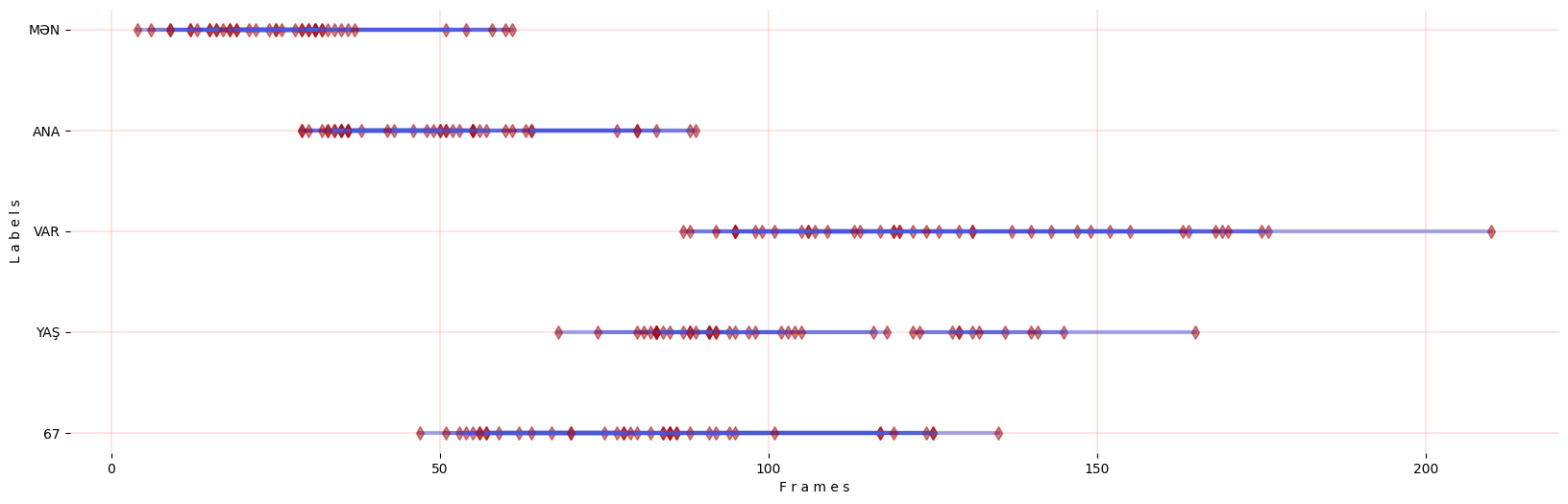}
\caption{Isolated word distribution within videos of a sentence}
\label{fig-ann}
\end{figure}

Fig. \ref{fig-ann} shows the variety of annotations throughout the frame range for one sentence. This sentence, comprising five labels (words), is represented by 21 different video samples and annotation files. The timing of sign appearances varies between samples of the same class while maintaining the same label order. Such labeling is beneficial for sign-boundary detection and enhances model robustness concerning temporal features \cite{star}.

Collectively, these annotations enhance the versatility of the sign-language video dataset. We provide a meticulously annotated dataset with frame-number-aligned JSON format annotation files and metadata containing spoken Azerbaijani translations for each sentence. Access to both sentence-level and word-level annotations enables researchers to develop models and algorithms that can accurately recognize individual words and translate entire sentences in sign language.

\subsection{Adherence to guidelines for SL AI datasets}

As we created the AzSL dataset for sign language recognition and translation purposes, we focused on finding the best practices for creating SL datasets for the same purpose. Different conventions for the notation of manual and non-manual activities were used in all the other SL AI datasets. The main common elements in creating a sign language dataset involve several steps including data collection, annotation, and documentation \cite{kopf_maria_2021_9561}. 

We hold in high regard the Fairness, Accountability, Transparency, and Ethics (FATE) \cite{10.1145/3436996} considerations in the creation of the AzSL dataset as well as the development of recognition and translation. \cite{10.1145/3436996} emphasizes the complexity of rights and responsibilities in data collection and storage, as well as the sensitivity of sign language usage by technologists due to the centrality of sign language to deaf cultural identity and a history of oppression \cite{desai2024systemicbiasessignlanguage}. We deliberated on the language included in the AzSL dataset, the diversity of the dataset, and its potential impact on marginalized communities. We ensured that it was inclusive, diverse, and transparent in creation and use. The vocabulary created for recordings is rich and versatile in terms of translation and does not include offensive speech.

Below we list the FATE issues along with our reflections to them:

\begin{itemize}
\item \emph{Content.} The scope of the vocabulary covers everyday speech and the Deaf service (social service) lexicon. AzSL fingerspelling, numbers (0-9 and greater), directions, date/time and other important glosses included. The sentences are deliberately composed as simple as possible and cover large, yet limited vocabulary.
\item \emph{Model Performance. } Demo project provided to see a case of model training.
\item \emph{Use Cases. } The intended use of the AzSL dataset is the application of computer vision techniques for isolated and continuous recognition as well as sign language translation.
\item \emph{Ownership. } The AzSL dataset is open-source and all the contributors without an exception gave consent to share the records.
\item \emph{Access. } We share the AzSL dataset with other research teams for benchmarking purposes.
\item \emph{Collection Mechanism. } Two HD RGB cameras of 30 fps were set for records, for front and side view. A suitable laboratory environment with right lighting was built.
\item \emph{Transparency and Understanding. } The content of the videos in the sign language dataset was created by signing actors and social service workers who know AzSL and work in the Deaf service    . These individuals have expertise in sign language and are familiar with the linguistic nuances and cultural aspects of AzSL. To enhance the quality and accuracy of the dataset, annotations were added by the signing actors during the recording process. This iterative process of recording and annotation allows for the refinement and improvement of the dataset, ensuring that the trained system can effectively interpret and recognize the signs for the target group of people.
\end{itemize}

\subsection{Data loader for ease of training and testing}

Data loaders are a fundamental component in machine learning pipelines, serving as the bridge between raw data and model training. The lack of well-functioning data loaders hinders their usability for the machine learning community \cite{challenges2022}. Providing a data loader for new sign language datasets is essential for the research community, particularly for benchmarking purposes. We have investigated existing datasets as \cite{githubGitHubJeraldflowersAmericanSignLanguageTensorFlow, githubGitHubSignlanguageprocessingdatasets} and adopted all the best in them. We have developed a data loader for AzSLD that can be adapted for other sign language datasets as well. By addressing the unique challenges posed by the dataset and adhering to best practices, the provided data loader can significantly enhance the efficiency and effectiveness of machine learning models in this domain. This data loader ensures the uniformity of input data, enabling seamless integration with machine learning pipelines.

\subsubsection*{Frame Extraction and Preprocessing}
The data loader extracts a fixed number of frames ($n_{\text{frames}}$) from each video file using a specified temporal resolution controlled by a frame step parameter. The frames are resized and padded to a uniform resolution ($224 \times 224$ pixels by default) while retaining the central region of interest. To ensure consistency across the dataset, the extracted frames are normalized and converted to a \texttt{float32} format. The frame extraction process handles videos of varying lengths by dynamically adjusting the starting point and padding with zero frames if required.

\subsubsection*{Label Encoding}
The loader supports label encoding by dynamically mapping categorical labels to one-hot encoded vectors. This is achieved through the creation of two dictionaries:
\begin{itemize}
    \item \textbf{Label-to-One-Hot Dictionary}: Maps each label to its corresponding one-hot encoded vector.
    \item \textbf{Index-to-Label Dictionary}: Provides a reverse mapping from the index of the one-hot vector to the original label.
\end{itemize}
These mappings are saved and loaded using the \texttt{pickle} library, enabling reusability across different runs.

\subsubsection*{Dataset Splitting}
The data loader incorporates a splitting function that divides the dataset into training and testing subsets. By default, 80\% of the data is allocated to the training set, with the remaining 20\% reserved for testing. This ensures a standardized pipeline for model evaluation and performance comparison.

\subsubsection*{Integration and Robustness}
The data loader is designed to handle diverse datasets with varying directory structures and video properties. Labels are extracted from directory names, enabling compatibility with hierarchically organized datasets. Furthermore, the implementation is robust to inconsistencies, such as videos with insufficient frames, ensuring that all data is formatted correctly for model training.

\subsubsection*{Benefits and Applications}
This data loader simplifies the preprocessing of sign language video datasets, making it suitable for various tasks such as gesture classification, action recognition, and translation systems. It ensures uniform data preparation, enhances reproducibility, and reduces the manual effort required to handle video data. This tool is a critical component of our pipeline, contributing to the efficient and accurate training of models for sign language recognition.

The source of the described data loader is publicly available at \cite{AzSLdataloader}.

\section{Limitations}

While the AzSLD dataset is a valuable resource for advancing research in sign language recognition and linguistics, it is important to acknowledge certain limitations. These constraints highlight areas for potential improvement and provide context for interpreting the dataset’s applicability in various research and development efforts. The following points outline the primary limitations:

\begin{itemize}
    \item \textbf{Limited Scope of Vocabulary:}  
    While the dataset includes a broad range of signs, it may not cover all possible vocabulary and expressions used in everyday communication or specialized contexts within the Azerbaijani Sign Language community. The absence of certain specialized or regional signs could limit its application in domain-specific or region-specific studies.
    
    \item \textbf{Controlled Recording Environment:}  
    The data collection was conducted in a controlled laboratory environment with a white background and fixed camera setups. This might not fully capture the variability and nuances of sign language usage in naturalistic settings. Consequently, models trained on this dataset may require additional adaptation or fine-tuning when applied to real-world environments.
    
    \item \textbf{Annotation Consistency:}  
    While the annotations are high-precision, manual annotation processes are inherently subject to human error and variability. Inter-annotator agreement metrics and consistency checks are essential but were not exhaustively detailed in this dataset. Future iterations could benefit from incorporating more robust validation processes to ensure uniformity across annotations.
    
    \item \textbf{Language Evolution:}  
    Sign languages evolve over time, and the dataset may not capture newly emerging signs or changes in usage patterns within the Azerbaijani Sign Language community. Regular updates to the dataset would be necessary to reflect these dynamic changes and keep it relevant for ongoing research.
\end{itemize}

By acknowledging these limitations, we aim to provide a transparent overview of the dataset's capabilities and encourage further efforts to address these challenges in future research and data collection initiatives. Moreover, collaborative contributions from the Azerbaijani Sign Language community could help in expanding the dataset's scope and improving its overall quality.

\subsection*{Acknowledgments}

The experiments were conducted at the ADA University's Center for Data Analytics Research (CeDAR). The data collection costs are covered by the ``Strengthening Data Analytics Research and Training Capacity through Establishment of dual Master of Science in Computer Science and Master of Science in Data Analytics (MSCS/DA) degree program at ADA University'' project, funded by BP and the Ministry of Education of the Republic of Azerbaijan.

Additionally, we would like to thank a number of people for contributing to the AzSL Dataset. Their participation in the data collection process has been instrumental in ensuring the authenticity and relevance of the Dataset. We acknowledge the following individuals for their knowledge, skills, and efforts in making this dataset possible: 
\begin{itemize}
\item Heydar Rahimli, Samir Sadigov, Kamran Abbasov from Azerbaijani Public Union "Support for the Deaf";
\item Aytaj Gadirova and Ziba Karimzade from "DOST" Agency for social protection in Azerbaijan;
\item AzSL interpreter: Yegana Taghiyeva;
\item ADA students: Ibrahim Alizada, Ismayil Shahaliyev, Habil Gadirli, Kamran Huseynov, Jalal Rasulzada, Rustam Safarli;
\item \parbox[t]{\linewidth}{Deaf users of AzSL: Yagub Naghiyev, Zahra Atakishiyeva, Ibrahim Amirov, Natig Mammadli, Farid Aghaverdiyev, Sama Taghiyeva, Elvina Orucova, Narinj Hashimova, 
Zahir Mammadov, Turkan Hasanli, Konul Azimova, Elshad Shirinov, Jamila Shamilova, Tural Mammadov, Elnara Orujova, Aziza Mirzayeva, Rovzat Abdulmabudov, Aliya Hashimova, Aysel Bakhshaliyeva, 
Havva Hasanova, Shamsi Shahbazova, Konul Abbasova, Murad Aliyev, Ali Aliyev, Fatima Huseynova, Konul Mammadli, Ramil Mammadov, Gabil Karimov, Afsana Karimova, Emin Huseynli, Damad Habibov, 
Parviz Jafarov, Nurlan Mohsumlu, Arzu Salimova, Ayisha Lukyan.}
\end{itemize}

\subsection*{Declarations}
The authors declare that they have no known competing financial interests or personal relationships that could have appeared to influence the work reported in this paper.

\section*{CRediT author statement}

\noindent
\textbf{Jamaladdin Hasanov:} Conceptualization, Methodology, Software, Data curation. \\
\textbf{Nigar Alishzade:} Software, Validation, Formal analysis, Investigation, Writing, Visualization.

\subsection*{Funding}
Not applicable

\section*{Data availability}
AzSLD is made available under the Creative Commons Attribution 4.0 International License on Zenodo repository at \url{https://zenodo.org/doi/10.5281/zenodo.13627300 }. Data loaders for utilizing this dataset are publicly accessible through the GitHub repository at \url{https://github.com/ADA-SITE-JML/azsl_dataloader}

\bibliographystyle{elsarticle-num}
\bibliography{references}

\end{document}